\documentclass[conference,final]{IEEEtran}

\pdfoutput=1

\IEEEoverridecommandlockouts
\IEEEpubid{\makebox[\columnwidth]{978-1-7281-1867-3/19/\$31.00~\copyright2019 IEEE \hfill} \hspace{\columnsep}\makebox[\columnwidth]{ }}

\usepackage{cite}
\usepackage{amsmath,amssymb,amsfonts}
\usepackage{graphicx}
\usepackage{textcomp}
\usepackage{xcolor}
\def\BibTeX{{\rm B\kern-.05em{\sc i\kern-.025em b}\kern-.08em
    T\kern-.1667em\lower.7ex\hbox{E}\kern-.125emX}}

\usepackage{float}
\usepackage{tabularx}
\usepackage{booktabs}
\usepackage{multirow}
\usepackage[ruled, vlined]{algorithm2e} 	

\makeatletter

\newcommand{\Rmnum}[1]{\expandafter\@slowromancap\romannumeral #1@}

\DeclareMathOperator*{\argmax}{arg\,max}

\makeatother

\begin{document}
\bstctlcite{IEEEexample:BSTcontrol}

\title{Statistical Linear Models in Virus Genomic Alignment-free Classification: Application to Hepatitis C Viruses
}

\author{\IEEEauthorblockN{Amine M. Remita}
\IEEEauthorblockA{\textit{Department of Computer Science} \\
\textit{Universit\'e du Qu\'ebec \`a Montr\'eal}\\
remita.amine@courrier.uqam.ca}
\and
\IEEEauthorblockN{Abdoulaye Banir\'e Diallo}
\IEEEauthorblockA{\textit{Department of Computer Science} \\
\textit{Universit\'e du Qu\'ebec \`a Montr\'eal}\\
diallo.abdoulaye@uqam.ca}
}

\maketitle

\begin{abstract}
Viral sequence classification is an important task in pathogen detection, epidemiological surveys and evolutionary studies. Statistical learning methods are widely used to classify and identify viral sequences in samples from environments. These methods face several challenges associated with the nature and properties of viral genomes such as recombination, mutation rate and diversity. Also, new generations of sequencing technologies rise other difficulties by generating massive amounts of fragmented sequences. While linear classifiers are often used to classify viruses, there is a lack of exploration of the accuracy space of existing models in the context of alignment free approaches.  In this study, we present an exhaustive assessment procedure exploring the power of linear classifiers in genotyping and subtyping  partial and complete genomes. It is applied to the Hepatitis C viruses (HCV). Several variables are considered in this investigation such as classifier types (generative and discriminative) and their hyper-parameters (smoothing value and regularization penalty function), the classification task (genotyping and subtyping), the length of the tested sequences (partial and complete) and the length of \textit{k}-mer words. Overall, several classifiers perform well given a set of precise combination of the experimental variables mentioned above. Finally,
we provide the procedure and benchmark data to allow for more robust assessment of classification from virus genomes. 
\end{abstract}


\section{Introduction}
\label{sec:intro}
Nucleotide sequence classification aims to assign an unlabeled or a new sequence (complete or partial) to a group of known sequences based on their characteristics. 
Sequence classification is used in several biomedical domains and comparative genomic fields such as pathogen detection \cite{Flygare2016, Ren2017}, taxonomic assignation of metagenomics reads \cite{Wang2007, Patil2012}, and epidemiological and evolutionary studies \cite{VanBelkum2001}. 
This task could be performed using an alignment-free approach which does not rely on building a multiple sequence alignment of known and unknown sequences. Such alignment-free approach could avoid some inconveniences since the sequence alignment can be time and resource consuming \cite{zielezinski2017}.
Alignment-free based approaches for sequence classification have shown promising results and performances compared to approaches based on alignment and phylogeny \cite{Bazinet2012, struck2014, zielezinski2017, remita2017, Lebatteux2019, Solis2018}. 

In previous works tackling taxonomic classification of metagenomics and viromics sequences, various tools implement statistical learning methods. 
For instance, in metagenomics, RDP \cite{Wang2007} and NBC \cite{Rosen2008} both implement a naive Bayes classifier. Also for taxonomic assignment \textit{PhyloPythiaS}+ implements structured output Support Vector Machines (SVM) framework \cite{Gregor2016}. A logistic regression model with L1 regularization was used in VirFinder, a tool for identifying viral sequences from metagenomic data \cite{Ren2017}.
For virus genome typing, a variable-order Markov model was implemented in COMET to classify Human Immunodeficiency Viruses 1 and 2 (HIV-1 and HIV-2, respectively) and Hepatitis C viruses (HCV)  \cite{struck2014}. SVM was used as a core model in CASTOR-KRFE method to extract a minimal set of features and classify viruses \cite{Lebatteux2019}. Other tools offer different types of classifiers depending on their best performances on a specific family of virus such as CASTOR \cite{remita2017} and KAMERIS \cite{Solis2018}.
In particular, some methods were developed and implemented in tools that perform genotyping and subtyping of HCV genomic sequences (for more details see Table A.I, available in the online supplemental material). 
Most of statistical learning classifiers previously mentioned are linear 
and can be classified into two categories: 1) Generative classifiers, which model the distribution of the input and output data (sequences and their taxonomic classes); 2) discriminative classifiers, which either model the posterior distribution of output data (taxonomic classes given a sequence) or find a discriminant function to map directly a sequence to its class \cite{bishop2006, hastie2009}.

In the context of alignment-free methods, composition representations of nucleotide sequences are widely used specially in sequence comparison and classification. They are based on the counts or frequencies of overlapping sub-sequences, with length \textit{k}, for a given sequence \cite{zielezinski2017}. These sub-sequences are known as words or \textit{k}-mers in the literature of sequence classification.


Here we assess the performance of generative and discriminative linear classifiers in genotyping and subtyping of partial and complete genomes of HCV. We examine the potential of these classifiers in classifying partial genomic sequences trained with complete genomes. Classification of genomic fragments is a challenge posed by current technologies of DNA sequencing that generate massive amounts of nucleotide fragments. We show that a global profile of \textit{k}-mer counts from complete genomes is sufficient to estimate the parameters of the models and classify correctly genomic fragments, and therefore no need to an explicit sampling of fragments in training setp. Several variables are considered in our assessment such as classifier types and their hyper-parameters, classification task, tested sequence lengths and \textit{k}-mer lengths.

This article is divided in three main section. We start by an overview of linear models that can be used in virus sequence classification in Section \ref{sec:linear}. Then we introduce the datasets and the benchmark procedure in Sections \ref{sec:hcv} and \ref{sec:setting} respectively. Section \ref{sec:results}  highlights the overall performance of each model and discusses the choice in choosing adequate experimental settings.

\section{Linear models for sequence classification}
\label{sec:linear}
\subsection{Sequence representation}
A nucleotide fragment $S$ is a sequence of $l \in \mathbb{N^*}$ ordered nucleotides $S = (s_1, s_2, ..., s_l) \in \mathcal{A}^l$, where $\mathcal{A} = \{A, C, G, T\}$ is the alphabet of nucleotides.
A $k$-mer word $u^a$ is a subsequence of length $k$ in $S$ at position $a$ such $u^a = S_{[a,\,a+k-1]}$ for $ a=1, 2, ..., d$ and $d=l-k+1$. 
Also, $S$ could be represented as a vector count $\textbf{x}$ of $m$ $k$-mer words $\textbf{x} = (x_1, x_2, ..., x_m)$, where $m = |\mathcal{A}|^k$ and $x_i$ is the number of occurrences of $k$-mer word $u_i$ in a sequence $S$, given by 
\begin{equation}
x_i = \sum_{j=1}^{d} \mathbb{I}(u_i, S_{[j,\,j+k-1]})
\end{equation}
and $\mathbb{I}(u_i, v) = 1$ if $u_i = v$, $0$ otherwise.
Thus, a dataset $D$ of $n$ nucleotide sequences will be represented by a $m \times n$ matrix $\textbf{X}  = (\textbf{x}_1, \textbf{x}_2, ..., \textbf{x}_n)$.\\
Finally, given a taxonomic rank (e.g. species or genotype), nucleotide sequences are labeled by a set of taxonomic disjoint classes $\mathcal{T} = \{T_1, T_2, ..., T_t\}, t \in \mathbb{N^*}$.

\begin{table*}[h]

\caption{Hepatitis C virus datasets from \cite{Lebatteux2019}.}

\begin{tabularx}{\textwidth}{ccccccc}
  \toprule
   Datasets & Group & Taxum & Avg Seq Len &  Classification & No. of  Instances [min-max] & No. of Classes  \\
   \midrule
   HCVGENCG& \Rmnum{4} ((+)ssRNA)& Hepatitis C virus & 9538 &  Genotypes & 284 [17-80] & 6  \\
   HCVSUBCG& \Rmnum{4} ((+)ssRNA)& Hepatitis C virus & 9538 &  Subtypes & 284 [4-25]& 18  \\
    \bottomrule
\end{tabularx}{}

\label{TAB1}
\end{table*}

\subsection{Linear classifiers}
In multiclass classification problems, a linear classifier $\mathcal{M}$ generates linear decision boundaries to separate instances of different classes \cite{hastie2009}. 
$\mathcal{M}$ is defined by a $m \times t$ matrix of weights $\textbf{W} = (\textbf{w}_1, \textbf{w}_2, ..., \textbf{w}_t)$ and an activation function $f(\cdot)$ which could be nonlinear. It has the form $f(\textbf{w}^\mathrm{T}\textbf{x} + w_0)$ \cite{bishop2006}.
$w_0$ represents the intercept of the model and $\textbf{w}_t = (w_{t_1}, w_{t_2}, ..., w_{t_m})$ the vector of weights for class $T_t$.
Learning classifiers differs in how they calculate the weights $\textbf{W}$ in order to optimize the classification and how to define the function $f(\cdot)$. 
To determine the class of a new sequence represented by a vector $\textbf{x}$, a classifier either models the posterior class probabilities $P(T_t\,|\,\textbf{x})$ and assigns the vector $\textbf{x}$ to a taxonomic class $\hat{T}$ that maximizes the posterior density as
\begin{equation}
    \label{eq:argmax1}
    \hat{T} = \argmax_t P(T_t\,|\,\textbf{x})
\end{equation}
or finds a \textit{discriminant} function to map directly the vector $\textbf{x}$ onto a class label \cite{bishop2006}.

Moreover, modeling the posterior class probabilities could be done using \textit{generative} or \textit{discriminative} approaches.

\subsubsection{Generative classifiers}
A generative approach models a joint probability density of the class and the sequence $P(T_t\,,\,\textbf{x})$ and uses Bayes' theorem (equation \ref{bayes}) to compute the posterior.
\begin{equation}
    \label{bayes}
    P(T_t\,|\,\textbf{x}) = \frac{P(T_t\,,\,\textbf{x})}{P(\textbf{x})} = \frac{P(\textbf{x}\,|\,T_t)\,P(T_t)}{P(\textbf{x})}
\end{equation}
With classifiers based on this approach, we can sample data from the modeled joint density $P(T_t\,,\,\textbf{x})$ in the space of vectors $\textbf{x}$.
Substituting equation \ref{bayes} in equation \ref{eq:argmax1} gives us:
\begin{equation}
    \label{eq:argmax2}
    \hat{T} = \argmax_t P(\textbf{x}\,|\,T_t)\,P(T_t)
\end{equation}
The probability density $P(\textbf{x})$ is constant over all classes and could be dropped in estimation of equation \ref{eq:argmax2}. The prior density of taxonomic classes $P(T_t)$ could be estimated using either a Maximum Likelihood Estimation (MLE) method or a Bayesian approach.
Several probabilistic classifiers adopt different approaches to model the class-conditional density $P(\textbf{x}\,|\,T_t)$. Here we provide details of  two types of generative classifiers: multinomial Bayes and Markov chain classifiers.

Multinomial Bayes (MB) classifiers model the class-conditional density by a multinomial distribution given by
\begin{equation}
    \label{eq_multinom}
    P(\textbf{x}\,|\,T_t) = \frac{d\,!}{\prod_{i=1}^{m} x_i\,!} \prod_{i=1}^{m}P(u_i\,|\,T_t)^{x_i}.
\end{equation}
Class-conditional probabilities of \textit{k}-mers $P(u_i\,|\,T_t)$ are the MB classifier parameters. Each parameter could be estimated by a MLE method or by a Bayesian approach (by adding a smoothing value $\alpha$).

Markov chain (Markov) classifiers model each nucleotide sequence as a ($k-1$)-order Markov chain model \cite{Durbin1998}.
\begin{equation}
    \begin{aligned}
    P(S\,|\,T_t) &= \prod_{i=k}^{l} P(s_i\,|\,S_{[i-k+1,\,i-1]},\,T_t)\\
         &= \prod_{i=k}^{l} \frac{P(u^{i-k+1}\,|\,T_t)}{P(v^{i-k+1}\,|\,T_t)},
    \end{aligned}
\end{equation}
where $u^{i-k+1}=S_{[i-k+1,\,i]}$ is a $k$-mer and $v^{i-k+1}=S_{[i-k+1,\,i-1]}$ is a $(k-1)$-mer at position $i-k+1$.
Therefore, the sequence $S$ will be represented by two vectors $\textbf{x}$ and $\textbf{z}$ corresponding to the profiles of words $w$ and $v$ respectively. The class-conditional density could be approximated by:
\begin{equation}
    \label{eq:markov}
    P(S\,|\,T_t) = P(\textbf{x},\textbf{z}|\,T_t) \approx \prod_{i=1}^{m} \frac{P(u_i\,|\,T_t)^{x_i}}{P(v_i\,|\,T_t)^{z_i}}.
\end{equation}
Similarly to a MB classifier, class-conditional probabilities of words $u_i$ and $v_i$ of Markov chain classifier could be determined by MLE or Bayesian methods.

\subsubsection{Discriminative classifiers}
A discriminative approach models directly the posterior density $P(T_t\,|\,\textbf{x})$ without assuming any distribution $P(\textbf{x}\,|\,T_t)$ for the input data.
Hence, a binary logistic regression (LR) models the posterior density using the logistic function which has the form:

\begin{equation}
    \begin{aligned}
     P(\mathcal{T} = t_0\,|\,\textbf{x}) &= \frac{exp(\textbf{w}^\mathrm{T}\textbf{x} + w_{0})} {1 + exp(\textbf{w}^\mathrm{T}\textbf{x} + w_{0})} \\
\textrm{and}\\
     P(\mathcal{T} = t_1\,|\,\textbf{x}) &= \frac{1} {1 + exp(\textbf{w}^\mathrm{T}\textbf{x} + w_{0})}.
    \end{aligned}
\end{equation}
Regularized LR fits the vector of parameters $\textbf{w}$ by joint minimization of the loss function and the regularization penalty function $R$ as follows:
\begin{equation}
    \min_\textbf{w} \,C\sum_{\textbf{x},t} log(1 + exp(-t(\textbf{w}^\mathrm{T}\textbf{x} + w_0))) + \lambda R(\textbf{w})
\end{equation}
where $t \in \{-1, 1\}$, C is a cost parameter and $\lambda$ is the regularization rate.
$R$ could take several forms such as L1 norm ($||\textbf{w}||_1 = \sum_i |w_i|$) or squared L2 norm ($||\textbf{w}||_2^2 = \sum_i w_i^2$).

Contrary to LR, linear support vector machine (LSVM) yields a discriminative function that maps input sequences to taxonomic classes. For soft-margin LSVM classifier, the optimal parameters $\textbf{w}$ are obtained by joint minimization of the hinge loss function or its squared and the penalty function as follows:
\begin{equation}
    \min_\textbf{w} \,C\sum_{\textbf{x},t} max(0, 1-t(\textbf{w}^\mathrm{T}\textbf{x} + w_0)) + \lambda R(\textbf{w}).
\end{equation}
LR and LSVM described here are binary classifiers. For multiclass taxonomic assignment we could use a one-versus-rest strategy where a classifier is learned per a taxonomic class against the other classes.

\section{Benchmark datasets: Hepatitis C viruses}
\label{sec:hcv}
Hepatitis C viruses (HCV) are an important cause of chronic liver disease and cancer \cite{giannini2003}. The World Health Organization has estimated that 71 million people have chronic hepatitis C infection and approximately $399\,000$ people die each year from hepatitis C \cite{WHO2019HCV}. HCV have a positive-sense single-stranded RNA genome of about $9600$ nucleotides. They belong to the Flaviviridae family of viruses.
HCV are classified into six confirmed genotypes with at least $30\%$ divergence among their genomes. At lower taxonomic level, genotypes are divided into several subtypes with $\sim20\%$ divergence \cite{simmonds2005}.
In this paper, we used two datasets of HCV that were constructed in our previous work \cite{Lebatteux2019}. 
Each dataset contains 284 complete genomes labeled by 6 genotypes in HCVGENCG dataset and 18 subtypes in HCVSUBCG dataset (Table \ref{TAB1}).

\section{Experimental setting}
\label{sec:setting}

We investigated the behavior and performance of generative and discriminative linear classifiers in gentoyping and subtyping HCV genomic sequences. Two cross-validation based strategies were devised to assess the abilities of both classifier types to classify whole-length (complete) and partial (fragment) genomes (described in Algorithm \ref{alg:cg} and Algorithm \ref{alg:ft}, respectively). For both strategies the classifiers were trained with complete genomes. 
Five-fold cross-validation is performed in all classification tasks. In each iteration the weighted F-measure was calculated on the test sequence data. Then, the overall performance of all iterations was averaged. The F-measure is given by the equation:
\begin{equation}
    \label{eq:fscore}
    \textrm{F-measure} = \frac{2 \times \textrm{Recall} \times \textrm{Precision}}{\textrm{Recall} + \textrm{Precision}}
\end{equation}
 
\begin{algorithm}[]
	\caption{Evaluation with complete genomes}
    \label{alg:cg}
	\SetKwInOut{Input}{Input}
	\SetKwInOut{Output}{Output}
	\SetKwBlock{Begin}{Begin}{End}
	\DontPrintSemicolon
	\Input{$D$: Complete genome sequences\\
	       $\mathcal{T}$: Respective taxonomic classes of $D$\\
	       $CLF$: Classifier with hyper-parameters\\
	}
	\Output{$F_{list}$ list of F-measure scores}
	\Begin
	{
	    \ForEach{$k$ $\in$ $[k_{min} ... k_{max}]$}
	    {
	        $\textbf{X} \gets build\_kmer\_matrix(D,\,k)$\\
	        Stratified split $\textbf{X}_D$ and $\mathcal{T}$ into $n$ folds\\
	        \ForEach{ fold $i$ $\in$ $[1 ... n]$}
	        {
	            \tcp{Build $\textbf{X}_{train}$ and $\textbf{X}_{test}$ from complete genomes}
	            
	            $\textbf{X}_{train} \gets \textbf{X} - \textbf{X}[i]$ ;
	            $\mathcal{T}_{train} \gets \mathcal{T} - \mathcal{T}[i]$\\
	            
	            $\textbf{X}_{test} \gets \textbf{X}[i]$ ;
	            $\mathcal{T}_{test} \gets \mathcal{T}[i]$\\
	            
	            \tcp{Learn and test model $\mathcal{M}$}
	            $\mathcal{M} \gets  learn\_model(CLF,\, \textbf{X}_{train}, \mathcal{T}_{train})$\\
	            $\mathcal{T}_{pred} \gets test\_model(\mathcal{M},\, \textbf{X}_{test})$\\
	            
	            $F_k \mathrel{+}= compute\_f\_measure(\mathcal{T}_{test}, \mathcal{T}_{pred}) / n$
	        }
	        $append(F_{list},\,F_k)$
	    }
    }
\end{algorithm}

\begin{algorithm}[]
	\caption{Evaluation with genomic fragments}
    \label{alg:ft}
	\SetKwInOut{Input}{Input}
	\SetKwInOut{Output}{Output}
	\SetKwBlock{Begin}{Begin}{End}
	\DontPrintSemicolon
	\Input{$D$: Complete genome sequences\\
	       $\mathcal{T}$: Respective taxonomic classes of $D$\\
	       $CLF$: Classifier with hyper-parameters\\
	       $ft\_size$: size of genomic fragments
	}
	\Output{$F_{list}$ list of F-measure scores}
	\Begin
	{
	    \ForEach{$k$ $\in$ $[k_{min} ... k_{max}]$}
	    {

	        Stratified and shuffled split $D$ and $\mathcal{T}$ into $n$ folds\\
	        \ForEach{ fold $i$ $\in$ $[1 ... n]$}
	        {
	            \tcp{Build $\textbf{X}_{train}$ from complete genomes}
	            $D_{train} \gets D - D[i]$ ;
	            $\mathcal{T}_{train} \gets \mathcal{T} - \mathcal{T}[i]$\\
	            $\textbf{X}_{train} \gets build\_kmer\_matrix(D_{train},\,k)$\\
	            
	            \tcp{Build $\textbf{X}_{test}$ from fragments}
	            $D_{test} \gets D[i]$ ;
	            $\mathcal{T}_{test} \gets \mathcal{T}[i]$\\
	            $G_{test}, \mathcal{T}_{G_{test}} \gets fragment\_genomes(D_{test},\,\mathcal{T}_{test}, \,ft\_size)$
	            $\textbf{X}_{test} \gets build\_kmer\_matrix(G_{test},\,k)$\\
	            
	            \tcp{Learn and test model $\mathcal{M}$}
	            $\mathcal{M} \gets  learn\_model(CLF,\, \textbf{X}_{train},\,\mathcal{T}_{train})$\\
	            $\mathcal{T}_{pred} \gets test\_model(\mathcal{M},\, \textbf{X}_{test})$\\
	            
	            $F_k \mathrel{+}= compute\_f\_measure(\mathcal{T}_{G_{test}}, \mathcal{T}_{pred}) / n$
	        }
	        $append(F_{list},\,F_k)$
	    }
    }
\end{algorithm}

\vspace{-0.5cm} 
\section{Results and discussion}
\label{sec:results}
Here, we present the experimental results when classifiers are trained on complete genomes but tested either with complete geneomes or genomic fragments. Then we highlight the overall observations that could drive subsequent studies.

\subsection{Evaluation with complete genomes}
The classifier performances were assessed in genotyping and subtyping of HCV complete genomes. The evaluation was based on a cross-validation strategy as described in Algorithm \ref{alg:cg}. 
In subtyping data (HCVSUBCG) the class \textit{6f} is underrepresented since it contains only four sequences. To ensure a large coverage of subtypes, we opted to not discard this class and use a 4-fold cross-validation. At least, this class will be represented by one sequence for each training fold. The results are presented in function of \textit{k}-mer lengths (\textit{k}) from 4 to 15 nucleotides. 

\begin{figure*}[ht]
	\includegraphics[width=\hsize]{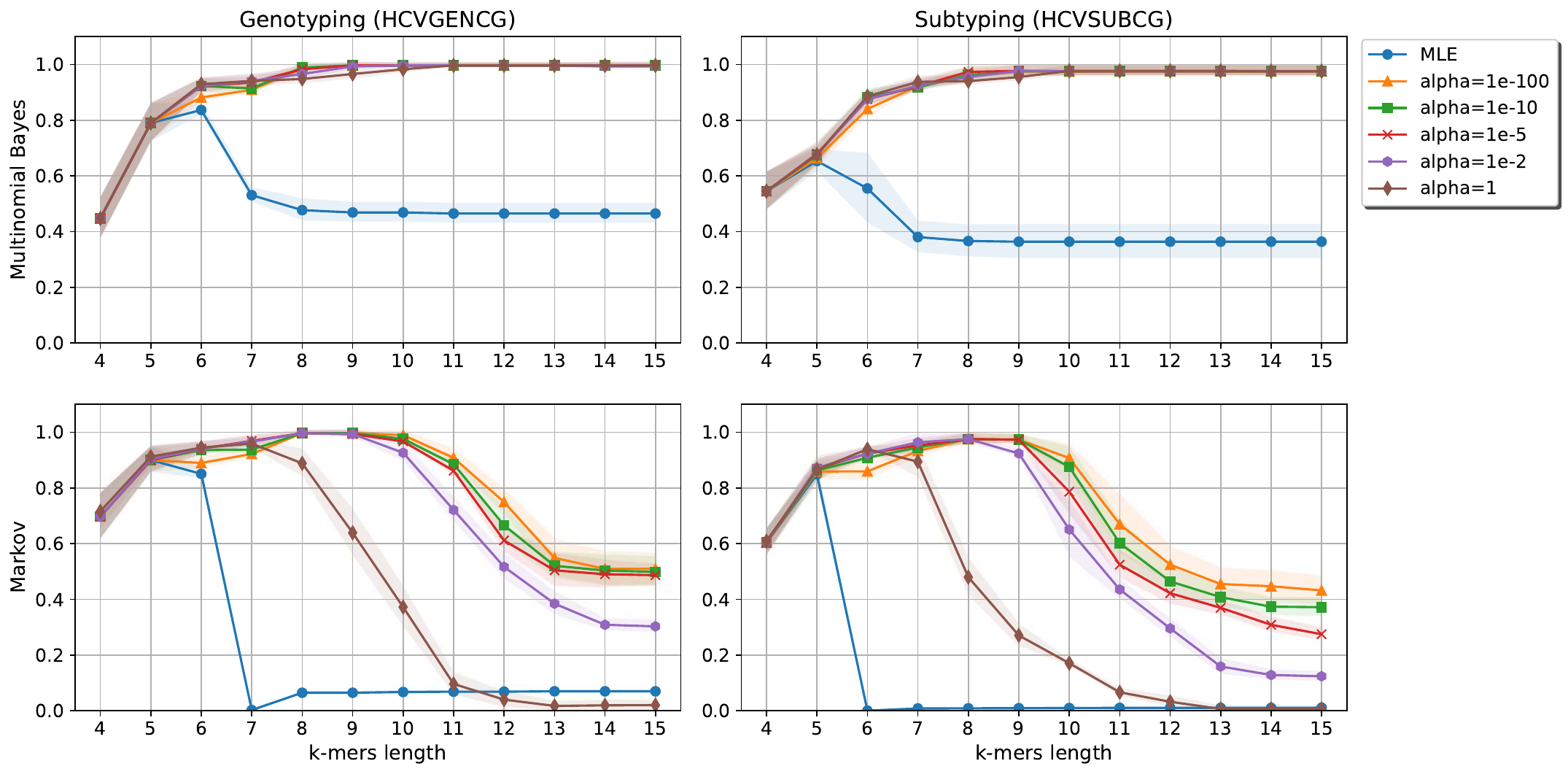}
	\caption{Averaged weighted F-measures of generative models tested on complete genomes. Filled regions correspond to the mean $\pm$ standard deviation of weighted F-measures of cross-validation iterations.}
	\label{fig:CG_GEN}
\end{figure*}

\begin{table*}[h]

\caption{Best and worst averaged weighted F-measures of linear models tested on complete genomes and their corresponding k lengths.}

\begin{tabularx}{\textwidth}{@{}llllllllll@{}}
\toprule
 &   & \multicolumn{4}{c}{Genotyping}  & \multicolumn{4}{c}{Subtyping}  \\ \cmidrule(l){3-10}
 &   & \multicolumn{2}{l}{Best} & \multicolumn{2}{l}{Worst} & \multicolumn{2}{l}{Best} & \multicolumn{2}{l}{Worst} \\ \cmidrule(l){3-10}
 Classifier & Model & F-measure & k lengths & F-measure  & k lengths  & F-measure & k lengths & F-measure & k lengths \\ \midrule
     \multirow{6}{*}{\begin{tabular}[c]{@{}l@{}}Multinomial \\ Bayes \end{tabular} } & MLE & 0.837 $\pm$ 0.019 & 6 & 0.448 $\pm$ 0.073 & 4  & 0.654 $\pm$ 0.039 & 5 & 0.364 $\pm$ 0.061 & 9-15 \\ 
 & alpha=1e-100 & 0.997 $\pm$ 0.007 & 9-15 & 0.448 $\pm$ 0.073 & 4  & 0.975 $\pm$ 0.019 & 11-15 & 0.545 $\pm$ 0.067 & 4 \\ 
 & alpha=1e-10 & 0.997 $\pm$ 0.007 & 9-15 & 0.448 $\pm$ 0.073 & 4  & 0.975 $\pm$ 0.019 & 9-15 & 0.545 $\pm$ 0.067 & 4 \\ 
 & alpha=1e-5 & 0.997 $\pm$ 0.007 & 9-15 & 0.448 $\pm$ 0.073 & 4  & 0.977 $\pm$ 0.017 & 9 & 0.545 $\pm$ 0.067 & 4 \\ 
 & alpha=1e-2 & 0.997 $\pm$ 0.007 & 10-13 & 0.448 $\pm$ 0.073 & 4  & 0.977 $\pm$ 0.017 & 9 & 0.545 $\pm$ 0.067 & 4 \\ 
 & alpha=1 & 0.997 $\pm$ 0.007 & 11-15 & 0.447 $\pm$ 0.073 & 4  & 0.977 $\pm$ 0.017 & 10-13 & 0.545 $\pm$ 0.067 & 4 \\  \midrule
 \multirow{6}{*}{Markov} & MLE & 0.900 $\pm$ 0.045 & 5 & 0.002 $\pm$ 0.003 & 7  & 0.849 $\pm$ 0.036 & 5 & 0.000 $\pm$ 0.001 & 6 \\ 
 & alpha=1e-100 & 0.997 $\pm$ 0.007 & 8-9 & 0.509 $\pm$ 0.060 & 14  & 0.975 $\pm$ 0.019 & 8 & 0.432 $\pm$ 0.051 & 15 \\ 
 & alpha=1e-10 & 0.997 $\pm$ 0.007 & 8-9 & 0.499 $\pm$ 0.054 & 15  & 0.975 $\pm$ 0.019 & 8 & 0.372 $\pm$ 0.034 & 15 \\ 
 & alpha=1e-5 & 0.997 $\pm$ 0.007 & 8 & 0.487 $\pm$ 0.038 & 15  & 0.975 $\pm$ 0.019 & 8 & 0.275 $\pm$ 0.023 & 15 \\ 
 & alpha=1e-2 & 0.997 $\pm$ 0.007 & 8 & 0.303 $\pm$ 0.022 & 15  & 0.975 $\pm$ 0.019 & 8 & 0.124 $\pm$ 0.017 & 15 \\ 
 & alpha=1 & 0.959 $\pm$ 0.018 & 7 & 0.017 $\pm$ 0.017 & 13  & 0.939 $\pm$ 0.007 & 6 & 0.007 $\pm$ 0.011 & 14 \\  \midrule
 \multirow{2}{*}{\begin{tabular}[c]{@{}l@{}}Logistic \\ Regression \end{tabular} } & LR\_L1 & 0.997 $\pm$ 0.007 & 4-11 & 0.996 $\pm$ 0.007 & 12-15  & 0.975 $\pm$ 0.019 & 4 & 0.941 $\pm$ 0.006 & 15 \\ 
 & LR\_L2 & 0.997 $\pm$ 0.007 & 4-15 & 0.997 $\pm$ 0.007 & 4-15  & 0.975 $\pm$ 0.019 & 4-5 & 0.967 $\pm$ 0.009 & 13-15 \\  \midrule
 \multirow{2}{*}{\begin{tabular}[c]{@{}l@{}}Linear \\ SVM \end{tabular} } & LSVM\_L1 & 1.000 $\pm$ 0.000 & 9-10 & 0.989 $\pm$ 0.015 & 13-14  & 0.975 $\pm$ 0.019 & 4 & 0.950 $\pm$ 0.018 & 12 \\ 
 & LSVM\_L2 & 0.997 $\pm$ 0.007 & 4-15 & 0.997 $\pm$ 0.007 & 4-15  & 0.975 $\pm$ 0.019 & 4-5 & 0.971 $\pm$ 0.018 & 13 \\ 

\bottomrule

\end{tabularx}

\label{TAB:CG}
\end{table*}

\subsubsection{Generative models}
In this study, we assessed the performance of two types of generative classifiers: multinomial Bayes (MB) and Markov chain (Markov) classifiers. 
In order to infer the parameters of these classifiers (class-conditional densities) we used Maximum Likelihood Estimation (MLE) and Bayesian approaches. For the Bayesian approach to estimating MB and Markov parameters we used different $\alpha$ values for smoothing: ($\alpha = 1e-100$, $1e-10$, $1e-5$, $1e-2$ or $1$).
The evaluation results of generative models on complete genomes are shown in Figure \ref{fig:CG_GEN} and Table \ref{TAB:CG}.

All MB models have almost the same performance in terms of weighted F-measure when $k\in\{4, 5\}$ (in genotyping from $0.447\pm0.072$ to $0.789\pm0.068$ and in subtyping from $0.545\pm0.06$ to $0.676\pm0.032$, respectively).
However, starting from $k=6$, the performance of MB models based on MLE (MLE-MB) was lower than those of models based on a Bayesian approach (B-MB). 
The weighted F-measure of MLE-MB model drops from its maximum value $0.837\pm0.019$ with $k=6$ to $0.465\pm0.035$ with $k\in[11,15]$ in genotyping and from $0.654\pm0.039$ with $k=5$ to $0.364\pm0.061$ with $k\in[9,15]$ in subtyping.
We noted an improvement in the performance of B-MB models when increasing \textit{k}-mer lengths either for genotyping or subtyping. 
Furthermore, with $k\in[11,13]$ in genotyping and with $k\in\{9,10\}$ in subtyping all B-MB models reach a maximum weighted F-measure of $0.997$.

Markov chain models based on MLE (MLE-Markov) show their best performance at $k=5$ with weighted F-measures of $0.900\pm0.045$ and $0.849\pm0.036$, in genotyping and subtyping respectively. 
At $k=7$ in genotyping and at $k=6$ in subtyping their performance drops drastically to reach an F-measure of $0.002\pm0.003$ and $0.000\pm0.001$, respectively and remain very low for longer \textit{k}-mers.
Unlike B-MB models, Markov models based on a Bayesian approach (B-Markov) do not maintain their performance high after reaching the maximum weighted F-measure values.
The maximum weighted F-measures occur at $k=8$ and equal to $0.997\pm0.007$ in genotyping and $0.975\pm0.019$ in subtyping for the majority of Markov models with smoothing. Moreover, in overall classification experiments with B-Markov models, smoothing with $\alpha=1$ shows the best performance with $k\in[4,6]$ but lower performance than the other Bayesian models when $k>7$.
However, $\alpha=1e-100$ smoothing shows the best performance with all lengths of \textit{k} except with $k\in[4,7]$.

\subsubsection{Discriminative models}
Discriminative models were represented by two classifiers in our work: logistic regression (LR) and linear support vector machine (LSVM). For both classifiers, we evaluated L1 and L2 penalties for regularization. The squared hinge loss function was used for LSVM classifier. Performance results on complete genomes are shown in Figures \ref{fig:hcv02} and A.1 for each model. 
In genotype taxonomic classification, LR and LSVM models classify the data with near perfect weighted F-measures ($>0.989\pm0.015$) across all \textit{k}-mer lengths. The best performance is shown by LSVM using L1-based regularization with $k\in[9,10]$ (weighted F-measure $=1.000\pm0.000$). In subtype classification, LR and LSVM model performances decrease slightly although the weighted F-measures remain greater than $0.941$ for all experiments (see Table \ref{TAB:CG}). The maximum weighted F-measure value of $0.975\pm0.019$ is reached by all discriminative models in subtyping. 
In general, L2-based models perform better than L1-based models which is clearly seen specially when $k>8$.

The evaluation with complete genomes shows that the overall performance varies substantially according to classifier types, their hyper-parameters and \textit{k}-mer lengths.
We observed different trend patterns of classification performances between generative and discriminative models, but same trends when comparing genotyping and subtyping tasks.
Most models could achieve a high weighted F-measure value of $0.997$ in genotyping and $0.939$ in subtyping except for MLE-based models. However, \textit{k}-mer lengths differ for these weighted F-measure values depending on each model instance. Using a SVM classifier, \cite{Lebatteux2019} reported quite similar results when classifying the same HCV genomic datasets (weighted F-measure $=1.000$ and $0.986$ in genotyping and subtyping, respectively \cite{Lebatteux2019}).


\subsection{Evaluation with genomic fragments}

\begin{figure*}[ht]
	\includegraphics[width=\hsize]{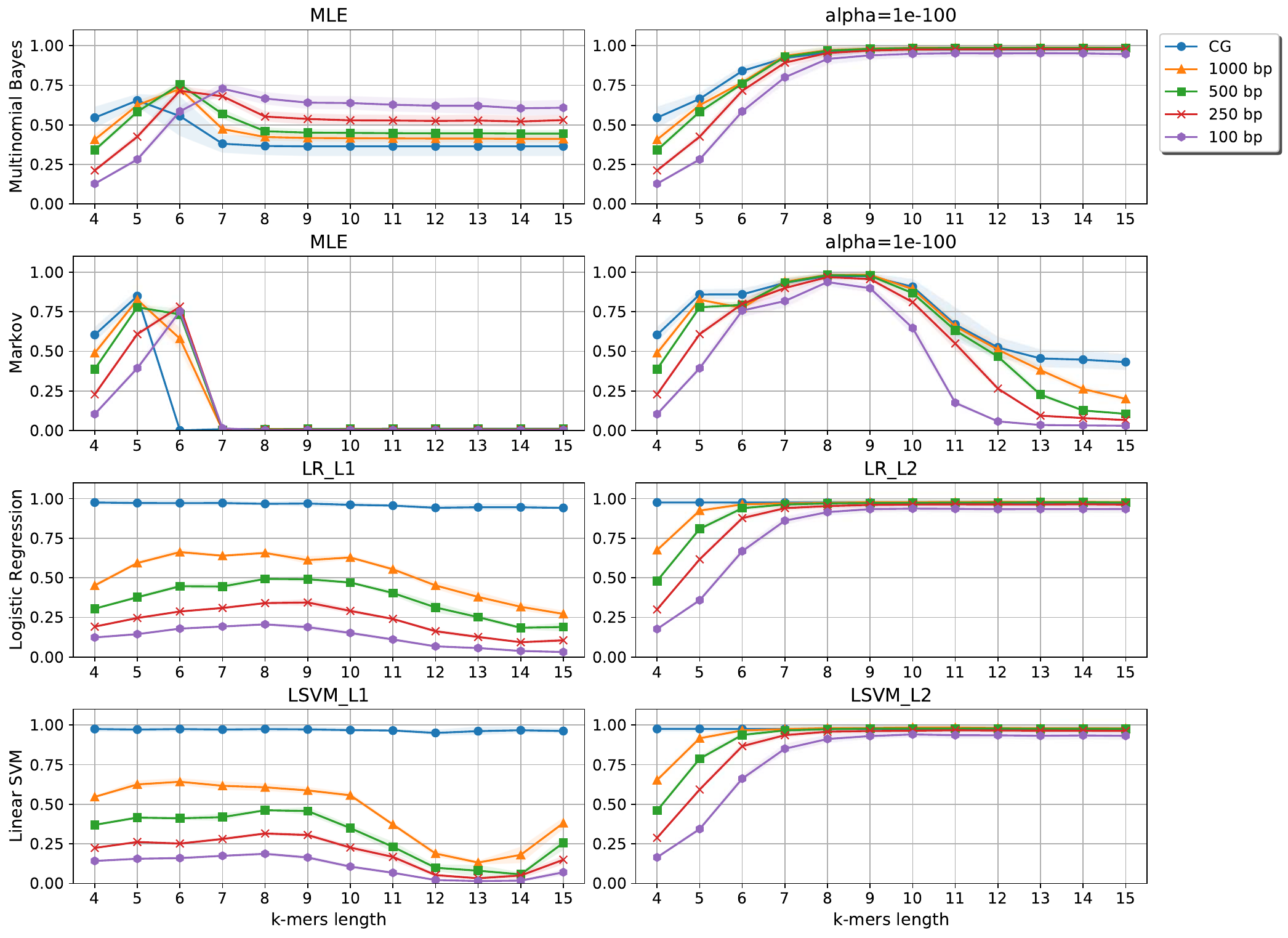}
	\caption{Averaged weighted F-measures of generative and discriminative models tested on different fragment lengths at subtyping (HCVSUBCG dataset). Filled regions correspond to the mean $\pm$ standard deviation of weighted F-measures of cross-validation iterations.}
	\label{fig:hcv02}
\end{figure*}

All proposed classifiers were trained with HCV complete genomes and tested with genomic fragments belonging to the same taxonomic classes as the genomes. 
We used a cross-validation strategy with 5-fold stratified and shuffled splits on the complete genome data. As described in Algorithm \ref{alg:ft}, in each iteration, a model is learned on a train set composed of complete genomes. After, it will be tested on fragmented sequences from the complete genomes of the test set.
Each class was sampled up to 1000 fragments. 
The classifiers were evaluated separately with fragments of lengths 100 bp, 250 bp, 500 bp and 1000 bp. The results of this evaluation are shown in Figure \ref{fig:hcv02}, and in Figure A.1 and Tables A.II-V available in the online supplemental material.

\subsubsection{Generative models}

With all fragment lengths, all MB models (including MLE-MB) have almost the same performance with \textit{k} lengths of $4$ and $5$ at genotyping and subtyping classification tasks. For these \textit{k}-mer lengths, the weighted F-measure ranges from $0.299\pm0.021$ to $0.857\pm0.019$ for genotyping and from $0.079\pm0.006$ to $0.829\pm0.017$ for subtyping (see Tables A.II-V). 
At $k=7$, the performance of MB models starts to diverge where we can note lower F-measure value for MLE-MB and B-MB with $\alpha=1e-100$ than the other models. After reaching its maximum, MLE-MB performance starts to drop at $k=7$ for all fragment lengths in genotyping.
However, the performance of B-MB models (including the model with $\alpha=1e-100$) increases after $k=8$ to exceed a weighted F-measure of $0.980$ at some values of $k$ for all fragment lengths and classification tasks.

Concerning Markov chain models, the performance of MLE-Markov attains its maximum at $k=6$ for all fragment lengths and both classification tasks except for testing on 500 bp and 1000 bp fragments at subtyping where the maximums were at $k=5$. The maximum values range from $0.769\pm0.023$ to $0.883\pm0.016$ at genotyping and from $0.752\pm0.040$ to $0.826\pm0.016$ at subtyping (see Tables A.II-V). 
After that, MLE-Markov performance on fragments falls and remains very low.
Comparing B-Markov models at genotyping the $\alpha=1e-100$-based model has consistently the maximal weighted F-measure values for all length fragments and with $k=9$ (values up to $0.988\pm0.016$).
Furthermore, in fragment subtyping, the performance of B-Markov models is highest at $k\in[6,8]$ depending on the smoothing value. The model with $\alpha=1e-5$ reaches the best weighted F-measure at $k=8$ with all fragment lengths for this classification task (values up to $0.987\pm0.013$). Subsequently, for both classification tasks, the performance drops gradually when $k>8$ to remain low with weighted F-measure $<0.472\pm0.068$.

\subsubsection{Discriminative models}
LR and LSVM models show similar behaviors when testing with different fragment lengths. 
As shown in Figures \ref{fig:hcv02} and A.1 and in Tables A.II-V, models with regularization penalty L2 perform better than those with penalty L1 independently of the classification variables (\textit{k}-mer lengths, fragment lengths and typing tasks).
For LR and LSVM L2-based models, the weighted F-measure exceeds a value of $0.900$ when \textit{k}-mers lengths are equal to or greater than $8$, $7$, $6$ and $5$ for 100 bp, 250 bp, 500 bp, 1000 bp fragments respectively. 
Furthermore, in both classification, the worst performance of L2-based models occurs when $k$ equals $4$ and fragment lengths is 100 bp (weighted F-measures $<0.332\pm0.011$). However, L2-based models keep a performance higher when $k$ is larger.
Models with L1 regularization penalty have lower performance and their weighted F-measure does not exceed $0.767\pm0.026$.
In fragment genotyping LR models with L1 penalty have highest weighted F-measure between $0.376\pm0.016$ and $0.767\pm0.026$ with $k=7$ for 100 bp fragments and $k=8$ for other fragment lengths. 
Moreover, in subtyping, they highlight a maximum weighted F-measure between $0.206\pm0.006$ and $0.662\pm0.017$ with $k=8$ for 100 and 500 bp, $k=9$ for 250 bp and $k=6$ for 1000 bp fragments. 
LSVM models regularized by L1 penalty have in general lower results compared to L1-based LR models. 
In fragment genotyping, maximum weighted F-measure for L1-based LSVM is between $0.323\pm0.007$ and $0.730\pm0.019$ with $k=7$ for 100 bp and 250 bp fragments, $k=8$ for 500 bp fragments and $k=6$ for 1000 bp fragments. 
In fragment subtyping, maximum performance is obtained with $k=8$ for all fragment lengths, except 1000 bp fragments($k=6$), and have weighted F-measure between $0.187\pm0.009$ and $0.642\pm0.021$. 
Unlike L2-based models, the capability of L1-based models in both classification declines when \textit{k}-mer lengths are longer.

\subsection{Overall remarks}
\label{sec:remarks}

We evaluated a set of linear classifiers in genotyping and subtyping of HCV genomic sequence represented by \textit{k}-mer profiles. This study evaluated various variables related to the level of taxonomic classification tasks (genotyping or subtyping), classifier types (generative or discriminative) and their hyper-parameters, genomic sequence lengths (complete or partial) and \textit{k}-mer lengths (from $4$ to $15$).

The results of this evaluation show that classification at low-level taxonomic clades is more difficult than at higher levels. In general, the classifiers performed better at genotyping than at subtyping HCV sequences. Several studies for viral \cite{remita2017, Lebatteux2019} and metagenomic \cite{Wang2007, Rosen2010, Liu2013} taxonomic classification have reported similar results where the performance is better at high-level classifications. Genomic sequences are more similar at low-level than at high-level clades, which makes more difficult to discriminate between sequences at low-level clades.

In both taxonomic levels, at least one or more generative and discriminative models reached a weighted F-measure $>0.950$ depending on their hyper-parameters and lengths of \textit{k}. Hence, we did not notice a clear advantage between both types of classifiers in term of weighted F-measure. However within each type, one model setting gave a clear advantage. For instance B-MB with $\alpha=1e-100$ and LSVM with L2 penalty are the best choice in generative and discriminative models respectively as they were stable among all experimental classifications.\\
As observed in previous studies \cite{Rosen2010, Liu2013, Vinje2015}, generative classifiers (MB and Markov) are sensitive to how they infer their parameters (class-conditional densities), either by MLE or Bayesian approaches. MLE approach could overfit and produce a sparse parameter matrix $\textbf{W}$ when unseen \textit{k}-mers in training step will have null estimates \cite{Rosen2010, Murphy2012} or very small probabilities will underflow the numerical precision \cite{bishop2006}.
Bayesian-based models could avoid null estimates and numerical underflow by adding positive and non-zero values $\alpha$ (pseudo-counts) \cite{Wong2009}. Hence, the MLE approach leads to inferior results compared to the Bayesian approach except for short \textit{k}-mers ($4$ and $5$), where they have similar performance and it is expected that most \textit{k}-mers are seen in the dataset.
Moreover, when $k\in\{6,7\}$, Bayesian models (MB and Markov) with larger $\alpha$ perform better than with smaller $\alpha$. However, this trend reverses when \textit{k}-mers are longer than $7$ nucleotides. \cite{Liu2013} reported comparable results in classifying microbial 16S and fungal 28S rRNA sequences with MB classifiers. At $k=8$, models with $\alpha>0.1$ have lower accuracies than models with $\alpha<0.0001$ in classifying full-length rRNA sequences \cite{Liu2013}.\\
After all, the B-Markov model performance decreases when $k>10$ despite the parameter smoothing. Larger \textit{k} lengths produce more sparse \textit{k}-mer profiles and when estimating Markov model parameters the division of \textit{k}-mer by (\textit{k}-1)-mer probability densities in equation \ref{eq:markov} gives small values that could underflow the numerical precision.

Discriminative classifiers (LR and LSVM) classified HCV genotypes and subtypes almost perfectly with complete genomes represented by any length of \textit{k}-mers. LR and LSVM models have similar behavior when they implement the same regularization penalty across all classification experiments. This suggests that their loss functions (logistic loss for LR and squared hing loss for LSVM) converge towards comparable results. When evaluating different classifiers for HIV-1 genome subtyping, \cite{Solis2018} concluded that SVM-based classifiers, and logistic regression achieved the highest performances. The authors reported accuracies of $96.49\%$ and $95.32\%$ for LSVM and LR respectively at $k=6$ \cite{Solis2018}.\\
In our study, the form of regularization did not influence the performance of neither classifiers, whereas it played a crucial role in classifying genomic fragments when model parameters are learned with global profiles of \textit{k}-mers from complete genomes. Regularization with L1 penalty decreases substantially the performance of classifiers on partial genomes unlike with L2 penalty.
On one hand, linear classifiers with L1 regularization produce sparse parameter matrix $\textbf{W}$ but dense matrix with L2 regularization \cite{Murphy2012}. On the other hand, genomic fragments generate also sparse vectors $\textbf{X}$ since they are partial. Hence, with L1-based models, a disagreement between $\textbf{W}$ values and those of $\textbf{X}$ could easily happen if $\textbf{W}$ was not learned from the same distribution of $\textbf{X}$ as in our evaluation with fragments.

Evaluated classifiers have better performance when tested with genomic sequences homologous to the sequences used in the learning step. Although surprising, some generative and discriminative models trained with complete genomes perform very well when they classify partial sequences. We observed that longer fragments are classified better than shorter ones. This was also observed in previous studies in virus genomic \cite{struck2014} and metagenomic \cite{Rosen2008,Liu2013,Matsushita2014} taxonomic classifications. Longer fragments generate less sparse data vectors and provide more information for classification \cite{Liu2013}.
Moreover, maximum performance of the evaluated models needs longer \textit{k}-mers when fragments are shorter, as with MB classifier in \cite{Rosen2008} and LR in \cite{Ren2017}.

Within the HCV datasets, the choice of the \textit{k}-mer length depends on all classification variables including taxonomic classification task, classifier types, hyper-parameters and sequence lengths. Mostly \textit{k}-mers with lengths between $8$ and $10$ are good options for classification tasks since they maximize the chance of achieving an optimal performance. However this observation does not hold for MLE-based models where the best option will be $k\in[5,6]$. Previous works in virus typing and identification reported optimal \textit{k}-mer lengths between $6$ and $9$ \cite{struck2014, Ren2017, Solis2018, Lebatteux2019}. Moreover, studies in metagenomic taxonomic assignment showed that MB classifiers need $k\in[12,15]$ to achieve good performance \cite{Rosen2008, Rosen2010, Matsushita2014}.



\section{Conclusion}
\label{sec:conclusion}
In this paper we provide an exhaustive procedure to assess the classification performance of different discriminative and generative classifiers with complete and partial genomes. We apply it to a benchmark of HCV viruses for genotyping and subtyping using several \textit{k}-mer lengths. The results highlight that there is no leading classifiers to perform on different experimental settings. Thus, the exploration of adequate experimental settings is required to capture the best performance. For HCV genomic data, this experimental settings allow to capture the highest predictive performance and to compare to the state-of-the art tools. Furthermore, most models perform well in either fragment and complete genome predictions. However, the hyper-parameters for estimating the model parameters and the \textit{k}-mer lengths vary in order to approximate the optimal performance. This study will be generalized to other viruses and the framework will be released to allow reproducible and accurate experimental settings for virus classification. 


\section*{Acknowledgment}
We would like to thank Dylan Lebatteux, Golrokh Vitae and Hayda Almeida for helpful discussion.\\
This research was enabled in part by support provided by Calcul Qu\'ebec and Compute Canada. 
It has also been supported by the Natural Sciences and Engineering Research Council of Canada (NSERC), the Fonds de recherche du Qu\'ebec - Nature et technologies (FRQNT), G\'enome Qu\'ebec and Genome Canada for the grants to ABD. AMR is NSERC and FRQNT fellow.


\bibliographystyle{IEEEtran}
\bibliography{IEEEabrv,bibm_remita_2019}

\onecolumn
\appendix

\setcounter{figure}{0}
\renewcommand\thefigure{A.\arabic{figure}}

\setcounter{table}{0}
\renewcommand\thetable{A.\Roman{table}}

\begin{table}[h]

\caption{Software and methods for HCV typing}

\begin{tabularx}{\textwidth}{@{}lllll@{}}
\hline
 & Software & Method & Availability & Reference \\ \cline{2-5} 
\multirow{4}{*}{Alignment-based} & LANL HCVBlast & Blast against a database of HCV sequences & \begin{tabular}[c]{@{}l@{}}Web interface\\ https://hcv.lanl.gov\end{tabular} & \cite{Kuiken2004} \\ \cline{2-5} 
 & HCV Typing Tool & Construct a phylogenetic tree & \begin{tabular}[c]{@{}l@{}}Web interface\\ https://www.genomedetective.com/app/typingtool/hcv\end{tabular} & \cite{deOliveira2005}  \\ \cline{2-5} 
 & MuLDAS & \begin{tabular}[c]{@{}l@{}}Construct distance matrix \\ + Multidimensional scaling\\ + Linear discriminant analysis (LDA)\end{tabular} & \begin{tabular}[c]{@{}l@{}}Web interface\\ http://gsa.muldas.org/index.cgi\end{tabular} & \cite{kim2010}  \\ \cline{2-5} 
 & Qiu et al. (2009) & \begin{tabular}[c]{@{}l@{}}Construct Position Weight Matrix\\ + SVM or random forest\end{tabular} & Not available & \cite{Qiu2009}  \\ \hline
\multirow{5}{*}{Alignment-free} & COMET & \begin{tabular}[c]{@{}l@{}}K-mer profiles\\ + Variable-order Markov model\end{tabular} & \begin{tabular}[c]{@{}l@{}}Web interface\\ https://comet.lih.lu/index.php?cat=hcv\end{tabular} & \cite{struck2014}  \\ \cline{2-5} 
 & CASTOR & \begin{tabular}[c]{@{}l@{}}RFLP-based features\\ + Feature selection\\+ Machine learning classifiers\end{tabular} & \begin{tabular}[c]{@{}l@{}}Web interface\\ http://castor.bioinfo.uqam.ca\end{tabular} & \cite{remita2017}  \\ \cline{2-5} 
 & KAMERIS & \begin{tabular}[c]{@{}l@{}}K-mer profiles\\ + Machine learning classifiers\end{tabular} & \begin{tabular}[c]{@{}l@{}}Open source code (MIT license)\\ https://github.com/stephensolis/kameris.git\end{tabular} & \cite{Solis2018}  \\ \cline{2-5} 
 & CASTOR-KRFE & \begin{tabular}[c]{@{}l@{}}K-mer profiles\\ + SVM-RFE feature selection\\ + SVM classifier\end{tabular} & \begin{tabular}[c]{@{}l@{}}Open source code (MIT license)\\ https://github.com/bioinfoUQAM/CASTOR\_KRFE.git\end{tabular} & \cite{Lebatteux2019}  \\ \cline{2-5} 
 & VGDC & Deep convolutional neural network & \begin{tabular}[c]{@{}l@{}}Free source code\\ https://github.com/afabijanska/VGDC.git\end{tabular} & \cite{Fabijanska2019}  \\ \hline
\end{tabularx}{\\RFLP stands for restriction fragment length polymorphism}

\label{TAB:HCV_progs}
\end{table}

\begin{table*}[h]

\caption{Averaged weighted F-measures of linear models tested on fragments of length 100 bp and their corresponding $k$ lengths.}

\begin{tabularx}{\textwidth}{@{}llllllllll@{}}
\toprule
 &   & \multicolumn{4}{c}{Genotyping}  & \multicolumn{4}{c}{Subtyping}  \\ \cmidrule(l){3-10}
 &   & \multicolumn{2}{l}{Best} & \multicolumn{2}{l}{Worst} & \multicolumn{2}{l}{Best} & \multicolumn{2}{l}{Worst} \\ \cmidrule(l){3-10}
 Classifier  & Model & F-measure & k lengths & F-measure  & k lengths  & F-measure & k lengths & F-measure & k lengths \\ \midrule
     \multirow{6}{*}{\begin{tabular}[c]{@{}l@{}}Multinomial \\ Bayes \end{tabular} } & MLE & 0.749 $\pm$ 0.027 & 7 & 0.264 $\pm$ 0.007 & 4  & 0.728 $\pm$ 0.035 & 7 & 0.128 $\pm$ 0.004 & 4 \\ 
 & alpha=1e-100 & 0.982 $\pm$ 0.011 & 14 & 0.264 $\pm$ 0.007 & 4  & 0.952 $\pm$ 0.024 & 13 & 0.128 $\pm$ 0.004 & 4 \\ 
 & alpha=1e-10 & 0.982 $\pm$ 0.011 & 14 & 0.264 $\pm$ 0.007 & 4  & 0.956 $\pm$ 0.022 & 11 & 0.128 $\pm$ 0.004 & 4 \\ 
 & alpha=1e-5 & 0.981 $\pm$ 0.011 & 12 & 0.264 $\pm$ 0.007 & 4  & 0.957 $\pm$ 0.020 & 11 & 0.128 $\pm$ 0.004 & 4 \\ 
 & alpha=1e-2 & 0.975 $\pm$ 0.010 & 12 & 0.264 $\pm$ 0.007 & 4  & 0.955 $\pm$ 0.019 & 11 & 0.128 $\pm$ 0.004 & 4 \\ 
 & alpha=1 & 0.963 $\pm$ 0.008 & 12 & 0.264 $\pm$ 0.007 & 4  & 0.940 $\pm$ 0.021 & 13 & 0.128 $\pm$ 0.004 & 4 \\  \midrule
 \multirow{6}{*}{Markov} & MLE & 0.769 $\pm$ 0.023 & 6 & 0.016 $\pm$ 0.005 & 9  & 0.752 $\pm$ 0.040 & 6 & 0.001 $\pm$ 0.001 & 8 \\ 
 & alpha=1e-100 & 0.967 $\pm$ 0.014 & 9 & 0.074 $\pm$ 0.008 & 15  & 0.937 $\pm$ 0.029 & 8 & 0.030 $\pm$ 0.002 & 15 \\ 
 & alpha=1e-10 & 0.965 $\pm$ 0.014 & 9 & 0.073 $\pm$ 0.007 & 15  & 0.941 $\pm$ 0.026 & 8 & 0.030 $\pm$ 0.002 & 15 \\ 
 & alpha=1e-5 & 0.963 $\pm$ 0.017 & 9 & 0.073 $\pm$ 0.007 & 15  & 0.944 $\pm$ 0.024 & 8 & 0.030 $\pm$ 0.002 & 15 \\ 
 & alpha=1e-2 & 0.965 $\pm$ 0.016 & 8 & 0.066 $\pm$ 0.007 & 15  & 0.941 $\pm$ 0.024 & 8 & 0.029 $\pm$ 0.003 & 15 \\ 
 & alpha=1 & 0.891 $\pm$ 0.018 & 7 & 0.042 $\pm$ 0.006 & 15  & 0.735 $\pm$ 0.036 & 7 & 0.013 $\pm$ 0.001 & 14 \\  \midrule
 \multirow{2}{*}{\begin{tabular}[c]{@{}l@{}}Logistic \\ Regression \end{tabular} } & LR\_L1 & 0.376 $\pm$ 0.016 & 7 & 0.113 $\pm$ 0.009 & 15  & 0.206 $\pm$ 0.006 & 8 & 0.031 $\pm$ 0.006 & 15 \\ 
 & LR\_L2 & 0.960 $\pm$ 0.011 & 13 & 0.332 $\pm$ 0.011 & 4  & 0.936 $\pm$ 0.023 & 10 & 0.176 $\pm$ 0.006 & 4 \\  \midrule
 \multirow{2}{*}{\begin{tabular}[c]{@{}l@{}}Linear \\ SVM \end{tabular} } & LSVM\_L1 & 0.323 $\pm$ 0.007 & 7 & 0.071 $\pm$ 0.006 & 15  & 0.187 $\pm$ 0.009 & 8 & 0.015 $\pm$ 0.002 & 13 \\ 
 & LSVM\_L2 & 0.963 $\pm$ 0.007 & 12 & 0.308 $\pm$ 0.007 & 4  & 0.941 $\pm$ 0.021 & 10 & 0.165 $\pm$ 0.006 & 4 \\ 

\bottomrule

\end{tabularx}

\label{TAB:FT_100}
\end{table*}

\begin{table}[h]

\caption{Averaged weighted F-measures of linear models tested on fragments of length 250 bp and their corresponding $k$ lengths.}

\begin{tabularx}{\textwidth}{@{}llllllllll@{}}
\toprule
 &   & \multicolumn{4}{c}{Genotyping}  & \multicolumn{4}{c}{Subtyping}  \\ \cmidrule(l){3-10}
 &   & \multicolumn{2}{l}{Best} & \multicolumn{2}{l}{Worst} & \multicolumn{2}{l}{Best} & \multicolumn{2}{l}{Worst} \\ \cmidrule(l){3-10}
 Classifier  & Model & F-measure & k lengths & F-measure  & k lengths  & F-measure & k lengths & F-measure & k lengths \\ \midrule
     \multirow{6}{*}{\begin{tabular}[c]{@{}l@{}}Multinomial \\ Bayes \end{tabular} } & MLE & 0.804 $\pm$ 0.009 & 6 & 0.301 $\pm$ 0.019 & 4  & 0.715 $\pm$ 0.028 & 6 & 0.212 $\pm$ 0.012 & 4 \\ 
 & alpha=1e-100 & 0.990 $\pm$ 0.010 & 13 & 0.301 $\pm$ 0.019 & 4  & 0.978 $\pm$ 0.014 & 14 & 0.212 $\pm$ 0.012 & 4 \\ 
 & alpha=1e-10 & 0.989 $\pm$ 0.010 & 13 & 0.301 $\pm$ 0.019 & 4  & 0.980 $\pm$ 0.013 & 14 & 0.212 $\pm$ 0.012 & 4 \\ 
 & alpha=1e-5 & 0.988 $\pm$ 0.012 & 13 & 0.301 $\pm$ 0.019 & 4  & 0.980 $\pm$ 0.014 & 13 & 0.212 $\pm$ 0.012 & 4 \\ 
 & alpha=1e-2 & 0.985 $\pm$ 0.011 & 13 & 0.301 $\pm$ 0.019 & 4  & 0.977 $\pm$ 0.014 & 13 & 0.212 $\pm$ 0.012 & 4 \\ 
 & alpha=1 & 0.974 $\pm$ 0.015 & 14 & 0.301 $\pm$ 0.019 & 4  & 0.965 $\pm$ 0.016 & 15 & 0.211 $\pm$ 0.011 & 4 \\  \midrule
 \multirow{6}{*}{Markov} & MLE & 0.850 $\pm$ 0.012 & 6 & 0.010 $\pm$ 0.005 & 8  & 0.782 $\pm$ 0.028 & 6 & 0.002 $\pm$ 0.001 & 7 \\ 
 & alpha=1e-100 & 0.985 $\pm$ 0.014 & 9 & 0.232 $\pm$ 0.030 & 15  & 0.968 $\pm$ 0.017 & 8 & 0.066 $\pm$ 0.005 & 15 \\ 
 & alpha=1e-10 & 0.984 $\pm$ 0.013 & 9 & 0.230 $\pm$ 0.029 & 15  & 0.972 $\pm$ 0.016 & 8 & 0.067 $\pm$ 0.005 & 15 \\ 
 & alpha=1e-5 & 0.983 $\pm$ 0.013 & 9 & 0.193 $\pm$ 0.023 & 15  & 0.973 $\pm$ 0.015 & 8 & 0.065 $\pm$ 0.005 & 15 \\ 
 & alpha=1e-2 & 0.979 $\pm$ 0.015 & 8 & 0.127 $\pm$ 0.016 & 15  & 0.972 $\pm$ 0.016 & 8 & 0.053 $\pm$ 0.005 & 15 \\ 
 & alpha=1 & 0.930 $\pm$ 0.014 & 7 & 0.057 $\pm$ 0.006 & 15  & 0.840 $\pm$ 0.023 & 6 & 0.021 $\pm$ 0.003 & 15 \\  \midrule
 \multirow{2}{*}{\begin{tabular}[c]{@{}l@{}}Logistic \\ Regression \end{tabular} } & LR\_L1 & 0.489 $\pm$ 0.012 & 8 & 0.201 $\pm$ 0.009 & 15  & 0.344 $\pm$ 0.021 & 9 & 0.093 $\pm$ 0.003 & 14 \\ 
 & LR\_L2 & 0.973 $\pm$ 0.012 & 15 & 0.434 $\pm$ 0.016 & 4  & 0.964 $\pm$ 0.018 & 14 & 0.300 $\pm$ 0.002 & 4 \\  \midrule
 \multirow{2}{*}{\begin{tabular}[c]{@{}l@{}}Linear \\ SVM \end{tabular} } & LSVM\_L1 & 0.441 $\pm$ 0.018 & 7 & 0.089 $\pm$ 0.010 & 15  & 0.315 $\pm$ 0.010 & 8 & 0.033 $\pm$ 0.006 & 13 \\ 
 & LSVM\_L2 & 0.972 $\pm$ 0.009 & 15 & 0.422 $\pm$ 0.023 & 4  & 0.967 $\pm$ 0.014 & 11 & 0.288 $\pm$ 0.005 & 4 \\ 

\bottomrule

\end{tabularx}

\label{TAB:FT_250}
\end{table}

\begin{table}[h]

\caption{Averaged weighted F-measures of linear models tested on fragments of length 500 bp and their corresponding $k$ lengths.}

\begin{tabularx}{\textwidth}{@{}llllllllll@{}}
\toprule
 &   & \multicolumn{4}{c}{Genotyping}  & \multicolumn{4}{c}{Subtyping}  \\ \cmidrule(l){3-10}
 &   & \multicolumn{2}{l}{Best} & \multicolumn{2}{l}{Worst} & \multicolumn{2}{l}{Best} & \multicolumn{2}{l}{Worst} \\ \cmidrule(l){3-10}
 Classifier  & Model & F-measure & k lengths & F-measure  & k lengths  & F-measure & k lengths & F-measure & k lengths \\ \midrule
     \multirow{6}{*}{\begin{tabular}[c]{@{}l@{}}Multinomial \\ Bayes \end{tabular} } & MLE & 0.831 $\pm$ 0.011 & 6 & 0.358 $\pm$ 0.027 & 4  & 0.755 $\pm$ 0.022 & 6 & 0.340 $\pm$ 0.016 & 4 \\ 
 & alpha=1e-100 & 0.991 $\pm$ 0.010 & 14 & 0.358 $\pm$ 0.027 & 4  & 0.985 $\pm$ 0.014 & 12 & 0.340 $\pm$ 0.016 & 4 \\ 
 & alpha=1e-10 & 0.990 $\pm$ 0.011 & 14 & 0.358 $\pm$ 0.027 & 4  & 0.986 $\pm$ 0.013 & 12 & 0.340 $\pm$ 0.016 & 4 \\ 
 & alpha=1e-5 & 0.989 $\pm$ 0.014 & 11 & 0.358 $\pm$ 0.027 & 4  & 0.986 $\pm$ 0.013 & 13 & 0.340 $\pm$ 0.016 & 4 \\ 
 & alpha=1e-2 & 0.987 $\pm$ 0.014 & 11 & 0.358 $\pm$ 0.027 & 4  & 0.985 $\pm$ 0.012 & 14 & 0.340 $\pm$ 0.016 & 4 \\ 
 & alpha=1 & 0.982 $\pm$ 0.012 & 14 & 0.358 $\pm$ 0.027 & 4  & 0.975 $\pm$ 0.013 & 15 & 0.340 $\pm$ 0.015 & 4 \\  \midrule
 \multirow{6}{*}{Markov} & MLE & 0.867 $\pm$ 0.015 & 6 & 0.012 $\pm$ 0.007 & 8  & 0.778 $\pm$ 0.014 & 5 & 0.002 $\pm$ 0.001 & 7 \\ 
 & alpha=1e-100 & 0.987 $\pm$ 0.014 & 9 & 0.378 $\pm$ 0.050 & 15  & 0.981 $\pm$ 0.013 & 8 & 0.105 $\pm$ 0.003 & 15 \\ 
 & alpha=1e-10 & 0.985 $\pm$ 0.013 & 9 & 0.344 $\pm$ 0.047 & 15  & 0.984 $\pm$ 0.013 & 8 & 0.097 $\pm$ 0.003 & 15 \\ 
 & alpha=1e-5 & 0.985 $\pm$ 0.013 & 9 & 0.283 $\pm$ 0.036 & 15  & 0.985 $\pm$ 0.013 & 8 & 0.084 $\pm$ 0.003 & 15 \\ 
 & alpha=1e-2 & 0.982 $\pm$ 0.014 & 8 & 0.180 $\pm$ 0.021 & 15  & 0.983 $\pm$ 0.013 & 8 & 0.067 $\pm$ 0.003 & 15 \\ 
 & alpha=1 & 0.930 $\pm$ 0.018 & 7 & 0.063 $\pm$ 0.008 & 15  & 0.897 $\pm$ 0.020 & 6 & 0.029 $\pm$ 0.006 & 15 \\  \midrule
 \multirow{2}{*}{\begin{tabular}[c]{@{}l@{}}Logistic \\ Regression \end{tabular} } & LR\_L1 & 0.621 $\pm$ 0.009 & 8 & 0.296 $\pm$ 0.009 & 15  & 0.493 $\pm$ 0.009 & 8 & 0.184 $\pm$ 0.011 & 14 \\ 
 & LR\_L2 & 0.979 $\pm$ 0.011 & 8 & 0.558 $\pm$ 0.013 & 4  & 0.976 $\pm$ 0.014 & 13 & 0.480 $\pm$ 0.009 & 4 \\  \midrule
 \multirow{2}{*}{\begin{tabular}[c]{@{}l@{}}Linear \\ SVM \end{tabular} } & LSVM\_L1 & 0.587 $\pm$ 0.003 & 8 & 0.160 $\pm$ 0.031 & 15  & 0.462 $\pm$ 0.006 & 8 & 0.057 $\pm$ 0.014 & 14 \\ 
 & LSVM\_L2 & 0.982 $\pm$ 0.011 & 13 & 0.540 $\pm$ 0.011 & 4  & 0.978 $\pm$ 0.013 & 10 & 0.460 $\pm$ 0.011 & 4 \\ 

\bottomrule

\end{tabularx}

\label{TAB:FT_500}
\end{table}

\begin{table}[h]

\caption{Averaged weighted F-measures of linear models tested on fragments of length 1000 bp and their corresponding $k$ lengths.}

\begin{tabularx}{\textwidth}{@{}llllllllll@{}}
\toprule
 &   & \multicolumn{4}{c}{Genotyping}  & \multicolumn{4}{c}{Subtyping}  \\ \cmidrule(l){3-10}
 &   & \multicolumn{2}{l}{Best} & \multicolumn{2}{l}{Worst} & \multicolumn{2}{l}{Best} & \multicolumn{2}{l}{Worst} \\ \cmidrule(l){3-10}
 Classifier  & Model & F-measure & k lengths & F-measure  & k lengths  & F-measure & k lengths & F-measure & k lengths \\ \midrule
     \multirow{6}{*}{\begin{tabular}[c]{@{}l@{}}Multinomial \\ Bayes \end{tabular} } & MLE & 0.848 $\pm$ 0.012 & 6 & 0.417 $\pm$ 0.029 & 4  & 0.727 $\pm$ 0.019 & 6 & 0.407 $\pm$ 0.029 & 4 \\ 
 & alpha=1e-100 & 0.994 $\pm$ 0.009 & 15 & 0.417 $\pm$ 0.029 & 4  & 0.987 $\pm$ 0.013 & 14 & 0.407 $\pm$ 0.029 & 4 \\ 
 & alpha=1e-10 & 0.993 $\pm$ 0.010 & 13 & 0.417 $\pm$ 0.029 & 4  & 0.987 $\pm$ 0.012 & 11 & 0.407 $\pm$ 0.029 & 4 \\ 
 & alpha=1e-5 & 0.990 $\pm$ 0.015 & 14 & 0.417 $\pm$ 0.029 & 4  & 0.988 $\pm$ 0.012 & 11 & 0.407 $\pm$ 0.029 & 4 \\ 
 & alpha=1e-2 & 0.989 $\pm$ 0.015 & 11 & 0.417 $\pm$ 0.029 & 4  & 0.987 $\pm$ 0.011 & 10 & 0.407 $\pm$ 0.029 & 4 \\ 
 & alpha=1 & 0.986 $\pm$ 0.014 & 13 & 0.417 $\pm$ 0.030 & 4  & 0.984 $\pm$ 0.011 & 14 & 0.407 $\pm$ 0.028 & 4 \\  \midrule
 \multirow{6}{*}{Markov} & MLE & 0.883 $\pm$ 0.016 & 6 & 0.036 $\pm$ 0.013 & 8  & 0.826 $\pm$ 0.016 & 5 & 0.002 $\pm$ 0.001 & 7 \\ 
 & alpha=1e-100 & 0.988 $\pm$ 0.016 & 9 & 0.472 $\pm$ 0.068 & 15  & 0.983 $\pm$ 0.011 & 9 & 0.199 $\pm$ 0.015 & 15 \\ 
 & alpha=1e-10 & 0.987 $\pm$ 0.016 & 9 & 0.409 $\pm$ 0.059 & 15  & 0.986 $\pm$ 0.013 & 8 & 0.139 $\pm$ 0.013 & 15 \\ 
 & alpha=1e-5 & 0.987 $\pm$ 0.016 & 8 & 0.347 $\pm$ 0.051 & 15  & 0.987 $\pm$ 0.013 & 8 & 0.104 $\pm$ 0.011 & 15 \\ 
 & alpha=1e-2 & 0.984 $\pm$ 0.015 & 8 & 0.209 $\pm$ 0.035 & 15  & 0.987 $\pm$ 0.012 & 8 & 0.057 $\pm$ 0.008 & 15 \\ 
 & alpha=1 & 0.943 $\pm$ 0.014 & 7 & 0.044 $\pm$ 0.007 & 15  & 0.917 $\pm$ 0.022 & 6 & 0.014 $\pm$ 0.005 & 14 \\  \midrule
 \multirow{2}{*}{\begin{tabular}[c]{@{}l@{}}Logistic \\ Regression \end{tabular} } & LR\_L1 & 0.767 $\pm$ 0.026 & 8 & 0.411 $\pm$ 0.016 & 15  & 0.662 $\pm$ 0.017 & 6 & 0.272 $\pm$ 0.021 & 15 \\ 
 & LR\_L2 & 0.989 $\pm$ 0.008 & 14 & 0.695 $\pm$ 0.014 & 4  & 0.979 $\pm$ 0.014 & 12 & 0.675 $\pm$ 0.014 & 4 \\  \midrule
 \multirow{2}{*}{\begin{tabular}[c]{@{}l@{}}Linear \\ SVM \end{tabular} } & LSVM\_L1 & 0.730 $\pm$ 0.019 & 6 & 0.220 $\pm$ 0.027 & 15  & 0.642 $\pm$ 0.021 & 6 & 0.132 $\pm$ 0.013 & 13 \\ 
 & LSVM\_L2 & 0.990 $\pm$ 0.010 & 15 & 0.691 $\pm$ 0.011 & 4  & 0.985 $\pm$ 0.013 & 10 & 0.653 $\pm$ 0.015 & 4 \\ 

\bottomrule
\end{tabularx}

\label{TAB:FT_1000}
\end{table}

\begin{figure}[h]
    \centering
	\includegraphics[width=\hsize]{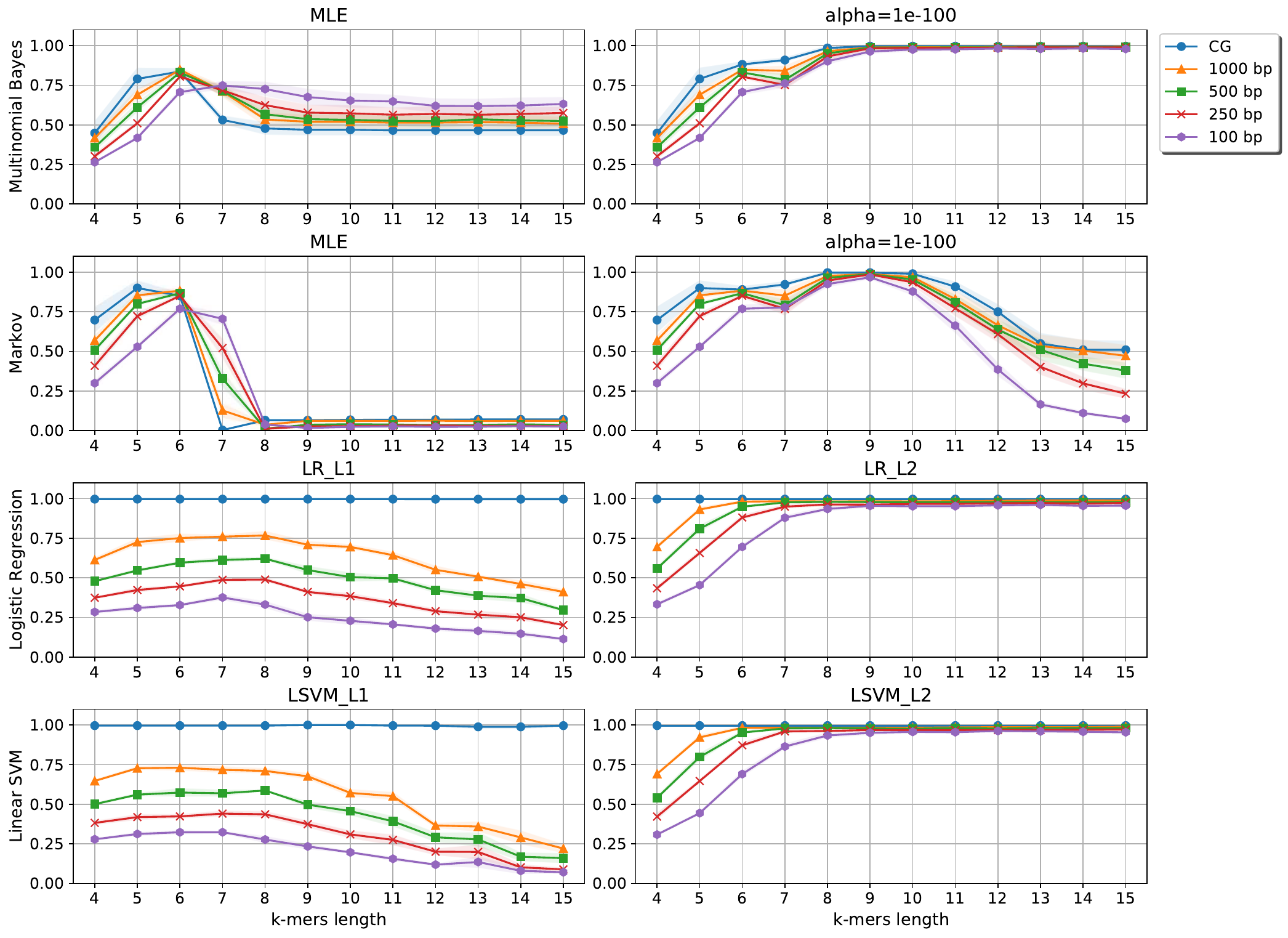}
	\caption{Averaged weighted F-measures of generative and discriminative models tested on different fragment lengths at genotyping (HCVGENCG dataset). Filled regions correspond to the mean $\pm$ standard deviation of weighted F-measures of cross-validation iterations.}
	\label{fig:hcv01}
\end{figure}

\end{document}